\documentclass[lettersize,journal]{IEEEtran}
\usepackage{amsmath,amsfonts}
\usepackage{algorithmic}
\usepackage{algorithm}
\usepackage{array}
\usepackage[caption=false,font=normalsize,labelfont=sf,textfont=sf]{subfig}
\usepackage{textcomp}
\usepackage{stfloats}
\usepackage{url}
\usepackage{verbatim}
\usepackage{graphicx}
\usepackage{cite}
\begin{document}

\title{TextMI: Textualize Multimodal Information
for Integrating Non-verbal Cues in Pre-trained Language Models}

\author {
    Md Kamrul Hasan\textsuperscript{\rm 1, *},
    Md Saiful Islam\textsuperscript{\rm 1},
    Sangwu Lee\textsuperscript{\rm 1},
    Wasifur Rahman\textsuperscript{\rm 1},\\
    Iftekhar Naim\textsuperscript{\rm 1},
    Mohammed Ibrahim Khan\textsuperscript{\rm 1},
    Ehsan Hoque\textsuperscript{\rm 1}\\

    \vspace{3mm}

    \small
    \textsuperscript{\rm 1} University of Rochester, United States\\

    \thanks{\textsuperscript{\rm *} Corresponding author; \textbf{Email:} hasankamrul@meta.com\\ 
    
    \copyright~2023 IEEE. Personal use of this material is permitted. Permission from IEEE must be obtained for all other uses, in any current or future media, including reprinting/republishing this material for advertising or promotional purposes, creating new collective works, for resale or redistribution to servers or lists, or reuse of any copyrighted component of this work in other works.}
}



\maketitle 

\begin{abstract}

Pre-trained large language models have recently achieved ground-breaking performance in a wide variety of language understanding tasks. However, the same model can not be applied to multimodal behavior understanding tasks (e.g., video sentiment/humor detection) unless non-verbal features (e.g., acoustic and visual) can be integrated with language. Jointly modeling multiple modalities significantly increases the model complexity, and makes the training process data-hungry.  While an enormous amount of text data is available via the web, collecting large-scale multimodal behavioral video datasets is extremely expensive, both in terms of time and money.  In this paper, we investigate whether large language models alone can successfully incorporate non-verbal information when they are presented in textual form. We present a way to convert the acoustic and visual information into corresponding textual descriptions and concatenate them with the spoken text. We feed this augmented input to a pre-trained BERT model and fine-tune it on three downstream multimodal tasks: sentiment, humor, and sarcasm detection. Our approach, TextMI, significantly reduces model complexity, adds interpretability to the model's decision, and can be applied for a diverse set of tasks while achieving superior (multimodal sarcasm detection) or near SOTA (multimodal sentiment analysis and multimodal humor detection) performance. We propose TextMI as a general, competitive baseline for multimodal behavioral analysis tasks, particularly in a low-resource setting.    

\end{abstract}

\section{Introduction}
\begin{figure*}[h]
\begin{center}

\includegraphics[width=\linewidth]{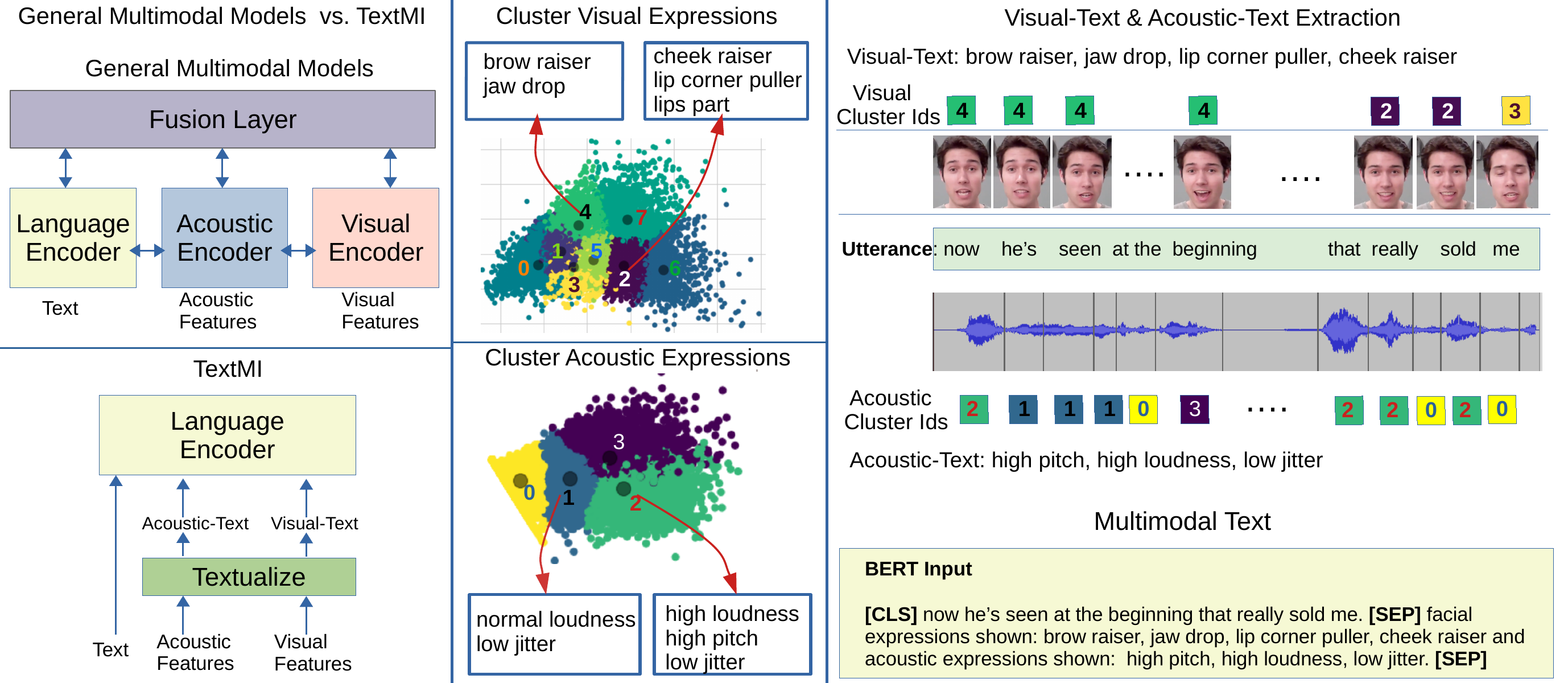}
\caption{Instead of using complex models that try to fuse multiple modalities, we rely on a single pre-trained language model (left). To textualize nonverbal (visual and acoustic) cues, we group unimodal features that frequently appear together into finite number of clusters, and then describe the clusters using text (middle). Finally, we extend the utterance text with the extracted visual-text and acoustic-text and pass it to a language model (right).}


\label{fig:nonverbal}
\end{center}
\end{figure*}
Humans are experts at understanding the nuances of non-verbal communication. To enable machines to understand those non-verbal signals, we often encode them in three modalities -- text, acoustic, visual -- and then fuse them using neural encoders. Recently, pre-trained large language models like BERT~\cite{devlin2018bert}  have become extremely effective in providing highly contextualized representation of the text modality. Acoustic and visual modalities, on the other hand, are typically converted into mid to high-level features  -- such as pitch, MFCC, facial expression,  etc. -- and then fed into transformers or LSTM encoders for fusion \cite{hazarika2020misa,han2021bi,hasan2021humor,Yu_Xu_Yuan_Wu_2021,han2021improving}. As a result, the encoders for acoustic and visual modalities are often trained from scratch compared to the pre-trained text encoders. Therefore, the fusion process is usually dominated by the highly contextualized text modality, making it difficult to infuse acoustic and visual information properly. Moreover, multimodal (text, acoustic, visual) behavioral datasets are typically smaller in size due to the higher cost of collecting them. The scarcity of large and diverse datasets makes it more challenging to train parameter-heavy multimodal models.


In contrast, the pre-trained models like BERT \cite{devlin2018bert} can be easily fine-tuned  to achieve state-of-the-art results for many NLP tasks like text-based humor detection \cite{mao2019bert} and sentiment analysis \cite{sun-etal-2019-utilizing}. 
In this paper, we ask -- can the language model understand non-verbal cues presented in text format? If we feed the language spoken and its associated nonverbal cues in textual format to a  pre-trained language model, can it analyze the multimodal content? How does it perform in learning human behavior compared to the existing resource-hungry multimodal models?

In this paper, we propose methods to textualize visual and acoustic cues present in multimodal (video) behavior understanding tasks: multimodal sentiment analysis, multimodal humor detection, and multimodal sarcasm detection.
We utilize widely-used tools such as Openface \cite{baltrusaitis2018openface} and Opensmile \cite{eyben2010opensmile} for extracting visual and acoustic features from videos. We then cluster these extracted visual and acoustic features into a finite number of groups using K-means clustering, as shown in Figure \ref{fig:nonverbal} (middle). We generate textual descriptions of these visual and acoustic clusters, and refer to them as \textit{visual-text} and \textit{acoustic-text} respectively. 
We prepare an extended text input by combining the text modality with the visual-text and acoustic-text (when available) as shown in Fig. \ref{fig:nonverbal} (right) and feed the extended text into a pre-trained BERT model. 

As \textit{multimodal} datasets, we use CMU-MOSI \cite{zadeh2016multimodal} and CMU-MOSEI~\cite{zadeh2018multimodal} for sentiment analysis, UR-FUNNY \cite{hasan-etal-2019-ur} for humor detection, and MUStARD \cite{castro2019towards} for sarcasm detection task. 
TextMI outperforms all state-of-the-art (SOTA) models in the multimodal sarcasm detection task and achieves near SOTA performance in the multimodal humor detection task. It also achieves better/near SOTA performances across all metrics in multimodal sentiment analysis task. 

These results demonstrate that our approach can be generalized across many diverse behavior-analysis tasks.
The proposed methodology, while fairly simple, can serve as a strong baseline for multimodal behavior understanding tasks, particularly with limited training data.

In summary, the main contributions of this paper are:
\begin{itemize}
    \item We propose a framework to convert visual and acoustic cues into natural language text for analyzing multimodal human behavioral tasks.
    \item We demonstrate that large language models can readily integrate visual and acoustic information provided in text format and achieve superior (or competitive) performance than the baseline multimodal models that use intricate fusion mechanisms. 
    \item TextMI is a simple and general methodology for multimodal behavior analysis that can act as a strong baseline for a diverse set of tasks. Our approach is \textit{interpretable} and particularly valuable for the tasks with limited data.

\end{itemize}

\section{Background}
In this section, we discuss some of the state-of-the-art multimodal models, their complexity and how the  scarcity of large multimodal (video) behavioral datasets makes the training challenging.


\subsection{Multimodal Models}

A lot of research exists on learning joint representation of multimodal data using LSTM, CNN, and fully-connected neural networks~\cite{zadeh2018memory,wang2018words,tsai2018learning,pham2018found,hazarika2018conversational,poria2017multi,liang2018multimodal,tsai2018learning,liu2018efficient,barezi2018modality}. We will focus Transformer-based models ~\cite{vaswani2017attention} which are the current state-of-the-art for modeling multimodal -- text,acoustic,visual-- modalities.  

Tsai et al. \cite{tsai2019multimodal} trained a model with a set of transformers: each transformer encoder learns its modality-specific encoding while interacting with the other encoders to capture the cross-modal interactions. Similarly, HKT~\cite{hasan2021humor} has modality-specific encoders for attending to each modality and a bimodal cross attention layer to jointly represent pairs of modality groups effectively. Several frameworks have been proposed where pre-trained \textbf{BERT} is used as language encoder and LSTM/Transformer based architectures are used to encode other modalities \cite{hazarika2020misa,han2021bi,Yu_Xu_Yuan_Wu_2021,han2021improving}. These architectures also introduced different intricate fusion mechanisms to achieve state-of-the-art results. Pham et al. ~\cite{pham2019found} translates from a source modality to a target modality while maintaining cycle consistency to ensure maximal retention of information in the joint representation. Another approach named Multimodal Routing~\cite{tsai2020multimodal} learns to dynamically adjust weights between input modalities and output vector for each data sample -- providing interpretation about cross-modality interactions for each example and whole dataset. In the domain of Graph-Neural-Networks, Model Temporal Attention Graph (MTAG) represents the sequential and temporal interactions in the multimodal sequence as nodes and edges ~\cite{yang2021mtag}.
These models have separate neural network components for capturing unimodal and cross-modal interactions that increase the number of parameters by the factor of the number of modalities. On the other hand, our approach uses a single pre-trained language model, and thus significantly reduces the model complexity and training time.

\subsection{Scarcity of Large Multimodal Behavioral Datasets}

Collecting large multimodal datasets for human behavioral tasks such as analyzing sentiment, job interview performance, disrespectful behavior, humor, etc. is extremely challenging, time-consuming, and expensive. The ratings for these tasks are often subtle and subjective and require expert knowledge and careful considerations. As a result, multimodal behavioral datasets are typically small and contain from hundreds to a few thousand training examples.

For example, two commonly used multimodal sentiment analysis datasets are CMU-MOSI~\cite{zadeh2016multimodal} and CMU-MOSEI ~\cite{zadeh2018multimodal}, consisting of 2199 and 23500 opinions respectively. ICT-MMMO dataset~\cite{wollmer2013youtube}, consisting of review videos, is also used for sentiment analysis with only 340 data instances. Similarly,
MOUD ~\cite{perez2013utterance} dataset contains 400 Spanish videos, each labeled with a positive, negative, or neutral sentiment. IEMOCAP~\cite{busso2008iemocap} dataset consists of 302 videos -- each annotated with the presence of 9 emotions as well as valence, arousal, and dominance. The UR-FUNNY dataset, the largest multimodal dataset for detecting humorous instances, contains 5k videos of humorous punchlines, with an equal number of negative samples~\cite{hasan-etal-2019-ur}. Multimodal sarcasm detection dataset \cite{castro2019towards} has only 690 video segments collected from several sitcoms. The scarcity of the large multimodal behavioral datasets makes it very challenging to train multimodal models with large number of parameters. In contrast, TextMI requires less parameters and easy to train on smaller datasets.


\section{Non-verbal Text Extraction}

In this section, we describe how we extract features from acoustic and visual modalities and convert them into corresponding text representations -- denoted as \textit{acoustic-text} and \textit{visual-text} (Figure \ref{fig:nonverbal}).

\subsection{Visual-text}
\label{ssec:visual_words}
OpenFace2 \cite{baltrusaitis2018openface} is used to extract facial Action Units (AU) features, based on the Facial Action Coding System (FACS) \cite{ekman1997face}. It is widely used in human affect analysis  ~\cite{hasan-etal-2019-ur,han2021bi,zadeh2017tensor}. Each of the action units represents a specific facial muscle movement. For example, AU02, AU04, and AU06 represent `Outer Brow Raiser', `Brow Lowerer', and `Cheek Raiser' respectively. These descriptions of the action units are readily understandable to humans. However, humans typically use a combination of these muscle movements simultaneously -- each denoted by an action unit --  to exhibit an expression. To mimic this behavior, we use K-means clustering to group the facial action units that co-occur often. In this paper, we use the following set of action units: [AU2, AU4, AU5, AU6, AU7, AU9, AU12, AU15, AU23, AU26, AU45].

First, we extract these eleven AU features from each frame of a video segment.  The timing information of each word is used to slice off the relevant range of action unit features for that word. Then, for each word, we average out the sliced visual feature array across the time dimension and get a visual feature vector (11 dimensions). Extracting word-aligned visual/acoustic vector is a common practice in multimodal behavior analysis tasks  \cite{chen2017multimodal,tsai2018learning,hazarika2020misa, han2021bi}. Once we have these visual feature vectors for each word across all videos, K-means clustering is used to group them into distinct sets. We use silhouette score to determine the optimal number of clusters. By analyzing the word-aligned visual features that belong to each cluster, we find the dominant (high intensity) action units in each cluster. Then, we represent each cluster by the text descriptions of the dominant action units. Table~\ref{tab:visword} shows the clusters, their dominant action units, and the corresponding descriptions of each action unit. These resulting textual descriptions are used to generate the visual-text.

Let, there are $n$ words in a video segment $U=[w_1,w_2,....,w_n]$. For the $i$th word ($w_i$), we can use the corresponding facial unit vector to extract the relevant cluster id. Thus, we can represent the cluster ids of the video utterance as: $C_v=[c_1,c_2,....,c_n]$. Each cluster-id represents a set of dominant AUs (e.g. table~\ref{tab:visword}). We sort all the AUs based on how many times they appear in the video utterance; the most commonly occurring ones are put at the beginning. For example, in Figure~\ref{fig:nonverbal}, the visual cluster id 4 has the highest frequency. This cluster id is represented by the dominant action units \textit{brow raiser} and \textit{jaw drop}. That is why these visual words appear at the beginning of the facial expression description. We concatenate all the visual words extracted from the sorted (based on frequency) AUs to generate the visual-text. We use k-means clustering since it gives a hard label to each of the clusters. Other techniques like the gaussian mixture model give the probabilities of a data point belonging to each of the K clusters -- making it difficult to convert these probabilities into words.

\begin{table}[t!]
\begin{center}
\resizebox{1\columnwidth}{!}{%
\begin{tabular}{|l| c | c |} 
 \hline
 \textbf{Cluster} &  \textbf{AUs} & \textbf{Textual Description (visual-text)}\\   \hline
 0 & None & neutral face \\ \hline
 1 & 15,25,20,26, & lip corner depressor, lip stretcher, jaw drop \\ 
 & 23,4,7& lip tightener, brow lowerer, lid tightener\\ \hline
 2 & 25,26 & jaw drop \\ \hline
 3 & 7,6,4,25,26 & lid tightener, cheek raiser, brow lowerer \\ \hline
 4 & 26,25,4 & jaw drop, brow lowerer, lip tightener \\ \hline 
 5 & 4,25 & brow lowerer, lips part \\ \hline
 6 & 12,6,25,7 & lip corner puller, cheek raiser, \\
 & &  lips part, lid tightener \\ \hline
\end{tabular}%
}
\end{center}
\caption{Visual-text for the clusters of visual features (from UR-FUNNY)}
\label{tab:visword}
\end{table}

\subsection{Acoustic-text}
Similar to extracting visual-text, we extracted the following interpretable acoustic features using Opensmile: pitch, loudness, jitter, and shimmer. Similar to the visual features, we extract word-aligned acoustic features of the whole dataset and apply K-Means clustering to them. Each cluster is assigned descriptions based on the intensity of the features present within it. A normal distribution is fitted to find the threshold of low, normal, and high intensity. Table~\ref{tab:acword} shows an example of the UR-FUNNY (multimodal humor detection) dataset. We denote these resulting descriptions as acoustic-text. 

For a video segment, first, we extract the cluster ids that are associated with the word-aligned acoustic vectors. Let, there are $n$ words in a video utterance $U=[w_1,w_2,....,w_n]$. Similar to the visual words, we extract the corresponding acoustic cluster ids $C_a=[c_1,c_2,....,c_n]$. Then, we create a textual description of the acoustic features by following the same methodology in \ref{ssec:visual_words}: replace each cluster-id with the underlying set of features (Table \ref{tab:acword}), sort the features by placing the most frequently appearing ones at the beginning, remove repeated features and concatenate all the texts.

\begin{table}[t!]
\begin{center}
\resizebox{1\columnwidth}{!}{%
\begin{tabular}{|l| c |} 
 \hline
 \textbf{Cluster} & \textbf{Textual Description (acoustic-text)}\\   \hline
 0 & high shimmer, high jitter, normal pitch, normal loudness \\ \hline
 1 & normal voice (every feature is in normal range) \\ \hline 
 2 & high loudness, high pitch, normal jitter, normal shimmer \\ \hline 
 3 & low pitch, low shimmer, normal loudness, normal jitter \\ \hline
\end{tabular}%
}
\end{center}
\caption{Acoustic-text extracted from UR-FUNNY}
\label{tab:acword}
\vspace{-4mm}
\end{table}

\subsection{Combining Text, Acoustic-text and Visual-text}
We append the acoustic-text and visual-text at the end of the text utterances separated by the separator token. As we experiment with the BERT language model, we represent the multimodal text as:
\texttt{[CLS] utterance text [SEP] Facial expressions shown: visual-text and acoustic expressions shown: acoustic-text [SEP]} (Figure~\ref{fig:nonverbal}).

\section{Experiments}
In this section, we discuss the datasets we use, the baseline models we compare with and the hyper parameter settings we experiment with. 

\subsection{Datasets}\label{sec:dataset}

\noindent \textbf{CMU-MOSI \& CMU-MOSEI:} Both the CMU-MOSI \cite{zadeh2016multimodal}  and the CMU-MOSEI \cite{zadeh2018multimodal}  are widely used benchmark datasets for evaluating a model's performance in predicting multimodal sentiment intensity. The CMU-MOSI is composed of 2199 video utterances segmented from 93 movie review videos. Each video utterance is manually annotated with a real number score ranging from -3 (most negative) to +3 (most positive) sentiment. The CMU-MOSEI dataset is an extension of the CMU-MOSI, but it increased the size of the dataset to 23,454 video utterances.


We use five different metrics following the previous works to evaluate the performance: mean absolute error (MAE), Pearson correlation, seven class accuracy (Acc-7), binary accuracy (Acc-2), and F1 score. Both Acc-2 and F1 score are computed for positive/negative (excluding zero).

\noindent \textbf{UR-FUNNY:} The UR-FUNNY \cite{hasan-etal-2019-ur} is a multimodal dataset of humor detection. It contains 10k video segments sampled from TED talks where punchline sentence is followed by context sentences. Each video segment is annotated with binary labels indicating if the punchline is humorous or not. 

\noindent \textbf{MUStARD:} Multimodal Sarcasm Detection Dataset \cite{castro2019towards} is compiled from popular TV shows like Friends, The Big Bang Theory, The Golden Girls, and Sarcasmaholics. 690 video segments are manually annotated with binary sarcastic/non-sarcastic labels. Each video segment has a target punchline sentence and the associated historical dialogues as context. Binary accuracy is used as the performance metric for both UR-FUNNY and MUStARD since both datasets have balanced test sets.

\subsection{Baseline Models}

Numerous methods have been proposed to learn the multimodal representation of text, acoustic and visual. We compare TextMI with the most recent and competitive baselines as mentioned below. 

\noindent\textbf{TFN} \cite{zadeh2017tensor} Tensor fusion network learns unimodal tensors and fuses them by three fold Cartesian product.

\noindent\textbf{LMF} \cite{liu2018efficient} creates multimodal fusion from the modality-specific low-rank factors by decomposing high dimensional tensors into many low dimensional factors.

\noindent\textbf{MFM} \cite{tsai2018learning} is a generative-discriminative model that factorizes representations into multimodal discriminative factors and modality-specific generative factors.

\noindent\textbf{ICCN}\cite{sun2020learning} learns the correlations between all three modalities of text, acoustic and visual via deep canonical correlation analysis.

\noindent\textbf{MulT} \cite{tsai2018learning} uses a set of transformer encoders to model inter-modal \& intra-modal interactions and combines their output in a late fusion manner.

\noindent\textbf{MISA} \cite{hazarika2020misa} projects all the video utterances into three modality-specific and one modality invariant spaces and then aggregates all those projections. 

\noindent\textbf{MAG-BERT} \cite{rahman2020integrating} introduced Multimodal Adaption Gate (MAG) to fuse acoustic and visual information in pretrained language transformers. During fine tuning, the MAG shifts the internal representations of BERT in the presence of the visual and acoustic modalities.

\noindent\textbf{BBFN } \cite{han2021bi} is an end-to-end network that performs fusion (relevance increment) and separation (difference increment) on pairwise modality representations.

\noindent\textbf{Self-MM} \cite{Yu_Xu_Yuan_Wu_2021} design a unimodal label generation strategy based on the self-supervised method that helps to learn modality specific representation. 

\noindent\textbf{MMIM } \cite{han2021improving} hierarchically maximizes the Mutual Information (MI) in multimodal fusion. It applies MI lower bounds for the unimodal inputs and the fusion stage.

\noindent\textbf{HKT} \cite{hasan2021humor} models multimodal humorous punchline using a set of transformer encoders and bi-modal cross attention layer. It also incorporates some humor-centric features extracted from external knowledge.

\textbf{State of the Art:} \textit{MMIM} is the state of the art model for the multimodal sentiment analysis task (in both CMU-MOSI and CMU-MOSEI datasets). Only MISA and HKT have experimented with the task of multimodal humor detection (UR-FUNNY) and multimodal sarcasm detection (MuSTARD) where \textit{HKT} has achieved SOTA performance.



\subsection{Experimental Design} Adam optimizer, linear scheduler with warmup and reduce learning rate at plateu scheduler are used to train the BERT language model. The search space of the learning rates is \{1e-05,3e-05,5e-05,1e-06\}. MSE loss is used to train models on CMU-MOSI and CMU-MOSEI as the sentiment intensity label is a real number between -3 to +3. Binary cross-entropy is used for other datasets. Dropout $[0.05-0.30]$ is used to regularize the model. All the experiments are run on K-80 \& A-100 GPUs. 

\vspace{6mm}

\begin{table*}[t!]
\begin{center}
\setlength\tabcolsep{3.2pt}
\resizebox{1.8\columnwidth}{!}{%
\begin{tabular}{ l| c  c c c  c | c  c c c  c  } 
 \hline
 Models & & CMU-MOSI & & &  &   & CMU-MOSEI & & & \\  
   & MAE  & Corr & Acc-7 & Acc-2 & F1 & MAE  & Corr & Acc-7 & Acc-2  & F1 \\ \hline
  TFN $^\diamond$ & 0.901 & 0.698 & 34.9 & 80.8 & 80.7 & 0.593 & 0.700 & 50.2 & 82.5 &82.1 \\ \hline
  LMF $^\diamond$ & 0.917 & 0.695 & 33.2 & 82.5 & 82.4 & 0.623 & 0.677 & 48.0 & 82.0 & 82.1 \\ \hline 
  MFM $^\diamond$&  0.877 &  0.706 & 35.4 & 81.7 & 81.6 & 0.568 & 0.717 & 51.3 & 84.4 & 84.3 \\ \hline 
  ICCN $^\diamond$& 0.860 & 0.710 & 39.0 & 83.0 & 83.0 & 0.565 & 0.713 & 51.6 & 84.2 & 84.2 \\ \hline
  MulT $^\ast\oplus$ & 0.832 & 0.745 & 40.1 & 83.3 & 82.9 & 0.570 & 0.758 & 51.1 & 84.5 & 84.5  \\ \hline
  MISA $^\diamond$ & 0.783 & 0.761 & 42.3 & 83.4 & 83.6 & 0.555 & 0.756 & 52.2 & 85.5 & 85.3 \\ \hline 
  MAG-BERT $^\ast\ddagger$ & 0.727 & 0.781 & 43.62 & 84.43 & 84.61 & 0.543 & 0.755 & 52.67 & 84.82 & 84.71\\ \hline
  BBFN  $^\oplus$& 0.776 & 0.755 & 45.0 & 84.3 & 84.3 & 0.529 & 0.767 & \underline{54.8} & \underline{86.2} & \underline{86.1} \\ \hline
  Self-MM $^\ast\ddagger$ & \underline{0.712} & \underline{0.795} & \underline{45.79} & 84.77 & 84.91 &  \underline{0.529} & 0.767 & 53.46 & 84.96 & 84.93\\ \hline
  MMIM (SOTA)$^\ddagger$ & \textbf{0.700} & \textbf{0.800} & \textbf{46.65} &  \textbf{86.06} & \textbf{85.98} & \textbf{0.526} & \underline{0.772} & \textbf{54.24} & 85.97 & 85.94 \\ \hline \hline 
  
  \textbf{TextMI} &  0.738  & 0.781 & 44.75 & \underline{85.21} & \underline{85.05} & 0.531 & \textbf{0.774}  & 53.21 & \textbf{86.45} & \textbf{86.38} \\ \hline

\end{tabular}%
}
\end{center}
\caption{Results of multimodal sentiment analysis task on CMU-MOSI and CMU-MOSEI datasets. The baseline results are taken from Hajarika et al.$^\diamond$ \cite{hazarika2020misa}, Han et al.$^\oplus$ \cite{han2021bi}, and Han et al. $^\ddagger$\cite{han2021improving}. Positive and negative samples (excluding zero) are used for calculating the Acc-2 and F1 score. MAE denotes Mean-absolute Error (lower is better), and Corr is Pearson Correlation (higher is better). The best performances are presented in \textbf{bold} and the closest competitors are \underline{underlined}. Some results ($^\ast$) were reproduced from the open source code and hyper parameter settings presented in the original paper. 
}
\label{tab:senti}
\end{table*}

\section{Results}
In this section, we present the performance of TextMI compared to the baseline multimodal models. We also report an ablation study to show the importance of having acoustic-text and visual-text alongside the main text.

\begin{table}[t!]
\begin{center}
\resizebox{0.65\columnwidth}{!}{%
\begin{tabular}{ l| c | c } 
 \hline
 Models &  UR-FUNNY & MUStARD \\  
  &  Acc & Acc \\ \hline
MISA $^\S$ & 70.61 & 66.18 \\ \hline 
HKT $^\S$ & \textbf{77.36} & 79.41\\ \hline
\textbf{TextMI} & 74.65 & \textbf{82.35} \\ \hline
\end{tabular}%
}
\end{center}
\caption{Results of multimodal humor (UR-FUNNY) and multimodal sarcasm (MUStARD) detection tasks. Since the datasets are balanced, accuracy for binary classification is reported, with best results highlighted in \textbf{bold}. The baseline results are taken from $^\S$\cite{hasan2021humor}.
}
\label{tab:humor}
\end{table}

\subsection{Multimodal Sentiment Analysis}
The results of multimodal sentiment analysis tasks are presented in Table~\ref{tab:senti}. TextMI achieves superior performance in the CMU-MOSEI dataset in terms of binary accuracy (0.48\% increase), F1 score (0.44\% increase), and pearson correlation (0.25\% relative improvement). Significance tests are run between TextMI and MMIM models for these metrics. We have run both models configured with the hyperparameters of best performances and changed the random seed only for different runs. Significant differences ($p<0.05$) are observed for these three metrics between TextMI and MMIM. For the other metrics, it is very close competitor of the SOTA model. TextMI also attain very competitive result in CMU-MOSI dataset across \emph{all} metrics. It achieves the second best performance in terms of binary accuracy and F1 score. These datasets are well studied for analyzing the performances of the multimodal models. All these baseline models use intricate fusion mechanism to summarize multimodal information. In contrast, our simple approach based on only pre-trained language encoder has achieved superior/near-SOTA performances. These results corroborate our hypothesis that the language model has the capability of understanding non-verbal cues presented in textual form.


\begin{figure}[t!]
\centering
\resizebox{0.49\textwidth}{!}{\includegraphics{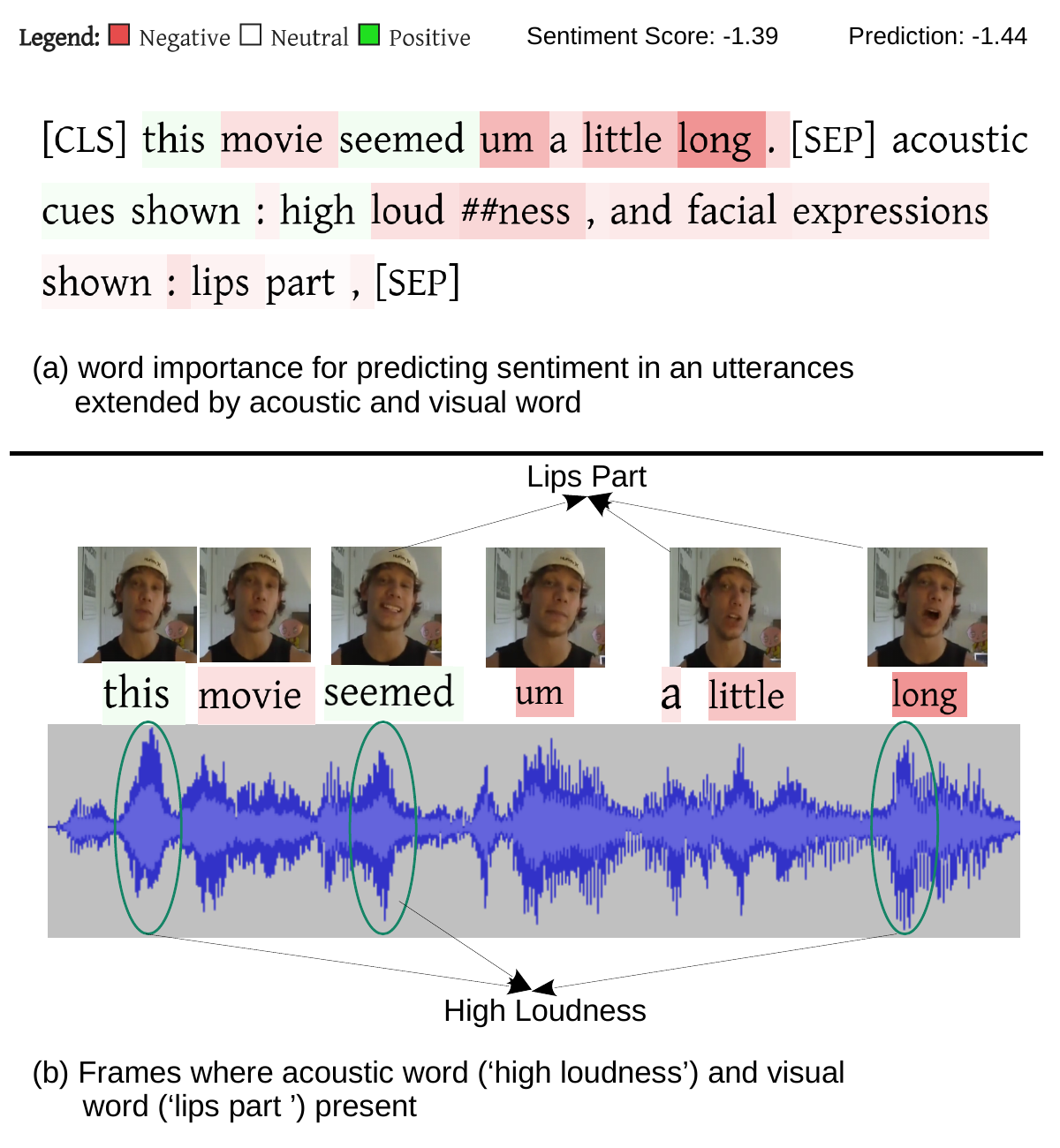}}
\caption{A multimodal sentiment analysis example to illustrate how the model put importance on text, acoustic and visual words. (a) word importance's are highlighted by color. (b) shows how the visual-text and acoustic-text are extracted. [CLS] and [SEP] are special tokens of BERT.}    
\label{fig:example}
\vspace{-4mm}
\end{figure}

\begin{figure*}[h]
\begin{center}

\includegraphics[width=\linewidth]{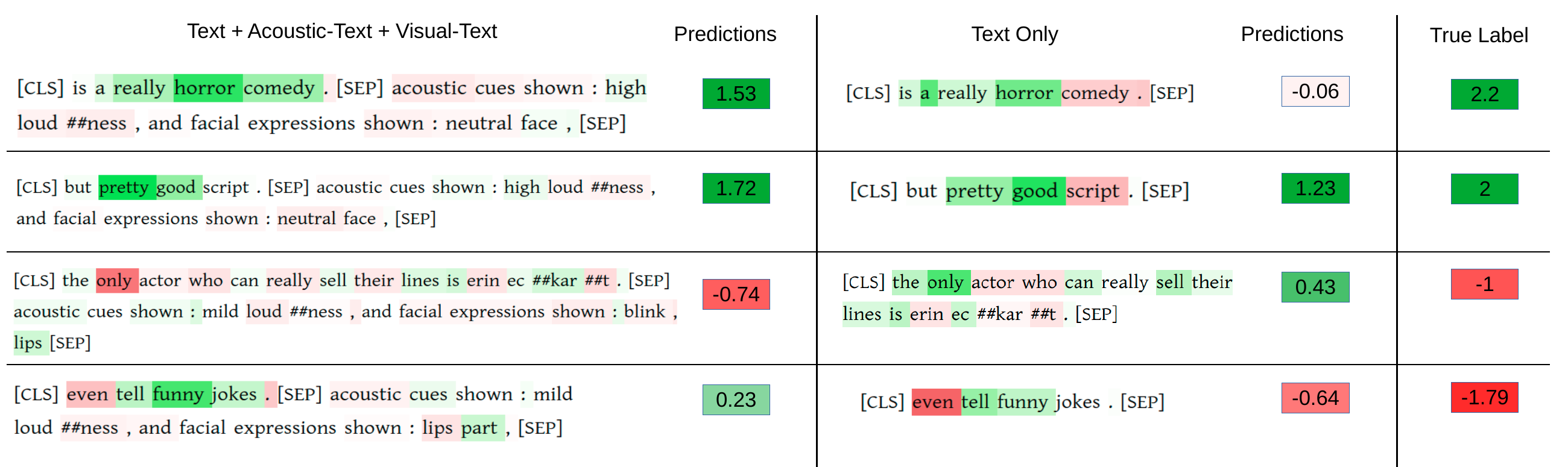}
\caption{Example from the CMU-MOSI dataset. The ground truth sentiment labels are between strongly negative
(-3) and strongly positive (+3). For each example, we show the Ground Truth and prediction output of both
the \textit{text+acoustic-text+visual-text} and \textit{text only}. Integrated gradients \cite{sundararajan2017axiomatic} method is used to decipher how each input token (across modalities) contributed to the model's final decision. [CLS] and [SEP] are special tokens of BERT.}
\label{fig:comparisn}
\end{center}
\end{figure*}

\subsection{Multimodal Humor and Sarcasm Detection}
The results of multimodal humor (UR-FUNNY) and sarcasm detection (MUStARD) tasks are presented in Table~\ref{tab:humor}. Since both datasets have a balanced test set, binary accuracy is reported as the performance metric. TextMI achieves superior performance (2.94\% increase) in the smaller MUStARD dataset (690 utterances) compared to the HKT. In comparison, HKT is superior in the relatively larger UR-FUNNY dataset (10K utterances). The HKT model incorporates external, task-specific features inspired by the theories of humor, and the availability of a large dataset helps train a complex model. However, the performance of TextMI that does not include any external task-specific feature is not far behind (2.71\% less accurate than HKT). This brings about an interesting perspective -- while generalizability of models is strongly desired, task-specific knowledge can be valuable and researchers might benefit from blending both general and task-specific knowledge into their models.


\subsection{Interpretation of Model Output}
The results discussed above indicate that incorporating non-verbal text into a language model has the capability of generalizing across various types of multimodal tasks. In addition, converting the facial and acoustic expressions into textual form makes it easy to interpret. Figure~\ref{fig:example} shows an example of multimodal sentiment analysis where input tokens are highlighted with colors based on the TextMI model's importance. Integrated gradients \cite{sundararajan2017axiomatic} method is used to decipher how each input token (across modalities) contributed to the model's final decision. We can see that the nonverbal-text such as high loudness plays an important role to identify the negative sentiment in this utterance. Since one of the major objectives of affective computing is understanding human emotion, such interpretability is highly valuable to the community.

\begin{table}[t!]
\begin{center}
\small
\resizebox{0.8\columnwidth}{!}{%
\begin{tabular}{ l| c | c | c | c   } 
 \hline
 Datasets & T  & T+V &  T+A  & T+A+V  \\   \hline
 CMU-MOSI & 83.99  & 84.76 & 84.45 & 85.21 \\ \hline 
 CMU-MOSEI & 85.76  & 86.15 & 86.20 & 86.45 \\ \hline 
 MUStARD & 79.41  & 79.41 & 80.88 & 82.35  \\ \hline
 UR-FUNNY & 74.04 & 74.65 & 74.45  & 74.65 \\ \hline
\end{tabular}%
}
\end{center}
\caption{Ablation study to show the importance of visual-text and acoustic-text. Binary accuracy is reported here as the performance metric. T= text, A= acoustic-text \& V= visual-text.}
\label{tab:abl}
\end{table}
\subsection{Role of Non-verbal Text}
To understand the roles of nonverbal-text quantitatively, we fine-tune the BERT encoder with text-only, text+acoustic-text and text+ visual-text information separately. The results are presented in Table~\ref{tab:abl}. Adding acoustic-text and visual-text improves the accuracy significantly, especially in the smaller datasets of CMU-MOSI and MUStARD. A qualitative analysis is presented in Figure~\ref{fig:comparisn}. In the first three examples, TextMI correctly predicted not only the polarity but also adjusted the sentiment intensity more accurately compared to the BERT model trained with text-only information. Additional information is present in visual-text and acoustic-text which text-only encoder could not utilize. The cross-attentions among the words of text utterance and nonverbal-text generate better scores. We also show an example where TextMI can fail (example 4). These examples and results demonstrate that the language model can integrate non-verbal cues (in text form) for affective computing tasks.


\section{Discussion and Future Work}

\textbf{Generalizability:} TextMI can be generalized across diverse multimodal tasks as the input is presented in textual format. Moreover, the use of pre-trained language models makes it easier to fine-tune on downstream tasks. We experimented with three multimodal behavior understanding tasks with four publicly available datasets. The superior results across these diverse tasks indicate the generalizability of TextMI. One limitation of our approach is that we depend on existing tools such as Openface and Opensmile. They may limit TextMI's performance since the error of these tools will propagate to the language model. Besides, if these tools are trained from biased data, it can hurt the fairness of our model as well. However, all the baseline multimodal models also have the same limitations as they used similar tools and features ~\cite{zadeh2017tensor,zadeh2018multimodal,tsai2018learning,hasan-etal-2019-ur,hazarika2020misa,han2021bi,han2021improving}. Building an end-to-end model trained from representative and unbiased data that can textualize the acoustic and visual features, and make inferences using a language model is a direction we plan to explore.\\

\noindent \textbf{Interpretability vs. Performance:} In a wide array of human behavior understanding tasks (e.g. identifying hateful speech video in social media),  it is of utmost importance to identify the key factors behind the model's decision. Since TextMI describes acoustic and visual information in textual format, it is much easier to interpret and visualize. Figure~\ref{fig:example} and Figure~\ref{fig:comparisn} illustrate how easy it is for humans to interpret the key acoustic and visual nuances when they are presented in textual format. However, as we concentrate on a set of interpretable features, our approach can result in loss of information and performance gained from incorporating complex features. 

Typically, visual/acoustic information is  modeled from scratch using low to mid-level features \cite{hazarika2020misa,han2021bi,hasan2021humor} through complex fusion processes. A complex multimodal model trained on a lot of data may be able to outperform our approach by sacrificing interpretability. However, our approach can be a very useful baseline for assessing models deployed in resource-constrained applications that must be interpretable.\\

\noindent \textbf{Dataset Size vs. Model's Complexity:} Unlike the existing multimodal models, TextMI uses only pre-trained language encoders, significantly reducing the model complexity and the number of parameters that are trained from scratch. Pre-trained large language models are generally sample efficient, and can learn well from fewer examples to solve downstream tasks. As a result, our model can be particularly useful for multimodal tasks with limited training data (CMU-MOSI \& MUStARD) -- making our approach very suitable for applying in new, exciting domains. A computationally efficient benchmark would make research of exciting problems more accessible since not all researchers have access to huge computing facilities.

\section{Conclusion}

In this paper, we show that large pre-trained language models can efficiently integrate non-verbal cues if they are described in textual form. Our approach achieves superior performance in multimodal sarcasm detection, and demonstrates better/near-SOTA performances in multimodal sentiment analysis and multimodal humor detection tasks, compared to the established baseline multimodal models that use intricate fusion mechanisms. An ablation study also indicates that the pre-trained language model performs better with acoustic and visual information than textual information only.
As our approach reduces the model's complexity and improves interpretability, it can be very useful for the scenarios where a large dataset is scarce or interpretability is requisite.
Though a multimodal model trained on a large dataset might provide better performance, our approach can still serve as a very strong and simpler baseline for future studies on multimodal behavioral tasks.

\bibliographystyle{IEEEtran}
\bibliography{references}

\begin{thebibliography}{10}
\providecommand{\url}[1]{#1}
\csname url@samestyle\endcsname
\providecommand{\newblock}{\relax}
\providecommand{\bibinfo}[2]{#2}
\providecommand{\BIBentrySTDinterwordspacing}{\spaceskip=0pt\relax}
\providecommand{\BIBentryALTinterwordstretchfactor}{4}
\providecommand{\BIBentryALTinterwordspacing}{\spaceskip=\fontdimen2\font plus
\BIBentryALTinterwordstretchfactor\fontdimen3\font minus
  \fontdimen4\font\relax}
\providecommand{\BIBforeignlanguage}[2]{{%
\expandafter\ifx\csname l@#1\endcsname\relax
\typeout{** WARNING: IEEEtran.bst: No hyphenation pattern has been}%
\typeout{** loaded for the language `#1'. Using the pattern for}%
\typeout{** the default language instead.}%
\else
\language=\csname l@#1\endcsname
\fi
#2}}
\providecommand{\BIBdecl}{\relax}
\BIBdecl

\bibitem{devlin2018bert}
J.~Devlin, M.-W. Chang, K.~Lee, and K.~Toutanova, ``Bert: Pre-training of deep
  bidirectional transformers for language understanding,'' \emph{arXiv preprint
  arXiv:1810.04805}, 2018.

\bibitem{hazarika2020misa}
D.~Hazarika, R.~Zimmermann, and S.~Poria, ``Misa: Modality-invariant
  and-specific representations for multimodal sentiment analysis,'' in
  \emph{Proceedings of the 28th ACM International Conference on Multimedia},
  2020, pp. 1122--1131.

\bibitem{han2021bi}
\BIBentryALTinterwordspacing
W.~Han, H.~Chen, A.~Gelbukh, A.~Zadeh, L.-p. Morency, and S.~Poria,
  ``Bi-bimodal modality fusion for correlation-controlled multimodal sentiment
  analysis,'' in \emph{Proceedings of the 2021 International Conference on
  Multimodal Interaction}, ser. ICMI '21.\hskip 1em plus 0.5em minus
  0.4em\relax New York, NY, USA: Association for Computing Machinery, 2021, p.
  6–15. [Online]. Available: \url{https://doi.org/10.1145/3462244.3479919}
\BIBentrySTDinterwordspacing

\bibitem{hasan2021humor}
M.~K. Hasan, S.~Lee, W.~Rahman, A.~Zadeh, R.~Mihalcea, L.-P. Morency, and
  E.~Hoque, ``Humor knowledge enriched transformer for understanding multimodal
  humor,'' in \emph{Proceedings of the AAAI Conference on Artificial
  Intelligence}, vol.~35, no.~14, 2021, pp. 12\,972--12\,980.

\bibitem{Yu_Xu_Yuan_Wu_2021}
\BIBentryALTinterwordspacing
W.~Yu, H.~Xu, Z.~Yuan, and J.~Wu, ``Learning modality-specific representations
  with self-supervised multi-task learning for multimodal sentiment analysis,''
  \emph{Proceedings of the AAAI Conference on Artificial Intelligence},
  vol.~35, no.~12, pp. 10\,790--10\,797, May 2021. [Online]. Available:
  \url{https://ojs.aaai.org/index.php/AAAI/article/view/17289}
\BIBentrySTDinterwordspacing

\bibitem{han2021improving}
W.~Han, H.~Chen, and S.~Poria, ``Improving multimodal fusion with hierarchical
  mutual information maximization for multimodal sentiment analysis,''
  \emph{Proceedings of the 2021 Conference on Empirical Methods in Natural
  Language Processing (EMNLP)}, 2021.

\bibitem{mao2019bert}
J.~Mao and W.~Liu, ``A bert-based approach for automatic humor detection and
  scoring.'' in \emph{IberLEF@ SEPLN}, 2019, pp. 197--202.

\bibitem{sun-etal-2019-utilizing}
C.~Sun, L.~Huang, and X.~Qiu, ``Utilizing {BERT} for aspect-based sentiment
  analysis via constructing auxiliary sentence,'' in \emph{Proceedings of the
  2019 Conference of the NAACL: Human Language Technologies}.\hskip 1em plus
  0.5em minus 0.4em\relax Association for Computational Linguistics, 2019, pp.
  380--385.

\bibitem{baltrusaitis2018openface}
T.~Baltrusaitis, A.~Zadeh, Y.~C. Lim, and L.-P. Morency, ``Openface 2.0: Facial
  behavior analysis toolkit,'' in \emph{2018 13th IEEE International Conference
  on Automatic Face \& Gesture Recognition (FG 2018)}.\hskip 1em plus 0.5em
  minus 0.4em\relax IEEE, 2018, pp. 59--66.

\bibitem{eyben2010opensmile}
F.~Eyben, M.~W{\"o}llmer, and B.~Schuller, ``Opensmile: the munich versatile
  and fast open-source audio feature extractor,'' in \emph{Proceedings of the
  18th ACM international conference on Multimedia}, 2010, pp. 1459--1462.

\bibitem{zadeh2016multimodal}
A.~Zadeh, R.~Zellers, E.~Pincus, and L.-P. Morency, ``Multimodal sentiment
  intensity analysis in videos: Facial gestures and verbal messages,''
  \emph{IEEE Intelligent Systems}, vol.~31, no.~6, pp. 82--88, 2016.

\bibitem{zadeh2018multimodal}
A.~B. Zadeh, P.~P. Liang, S.~Poria, E.~Cambria, and L.-P. Morency, ``Multimodal
  language analysis in the wild: Cmu-mosei dataset and interpretable dynamic
  fusion graph,'' in \emph{Proceedings of the 56th Annual Meeting of the
  Association for Computational Linguistics (Volume 1: Long Papers)}, vol.~1,
  2018, pp. 2236--2246.

\bibitem{hasan-etal-2019-ur}
\BIBentryALTinterwordspacing
M.~K. Hasan, W.~Rahman, A.~Bagher~Zadeh, J.~Zhong, M.~I. Tanveer, L.-P.
  Morency, and M.~E. Hoque, ``{UR}-{FUNNY}: A multimodal language dataset for
  understanding humor,'' in \emph{Proceedings of the 2019 Conference on
  Empirical Methods in Natural Language Processing and the 9th International
  Joint Conference on Natural Language Processing (EMNLP-IJCNLP)}.\hskip 1em
  plus 0.5em minus 0.4em\relax Hong Kong, China: Association for Computational
  Linguistics, Nov. 2019, pp. 2046--2056. [Online]. Available:
  \url{https://www.aclweb.org/anthology/D19-1211}
\BIBentrySTDinterwordspacing

\bibitem{castro2019towards}
S.~Castro, D.~Hazarika, V.~P{\'e}rez-Rosas, R.~Zimmermann, R.~Mihalcea, and
  S.~Poria, ``Towards multimodal sarcasm detection (an \_obviously\_ perfect
  paper),'' \emph{arXiv preprint arXiv:1906.01815}, 2019.

\bibitem{zadeh2018memory}
A.~Zadeh, P.~P. Liang, N.~Mazumder, S.~Poria, E.~Cambria, and L.-P. Morency,
  ``Memory fusion network for multi-view sequential learning,'' in
  \emph{Thirty-Second AAAI Conference on Artificial Intelligence}, 2018.

\bibitem{wang2018words}
Y.~Wang, Y.~Shen, Z.~Liu, P.~P. Liang, A.~Zadeh, and L.-P. Morency, ``Words can
  shift: Dynamically adjusting word representations using nonverbal
  behaviors,'' \emph{arXiv preprint arXiv:1811.09362}, 2019.

\bibitem{tsai2018learning}
Y.-H.~H. Tsai, P.~P. Liang, A.~Zadeh, L.-P. Morency, and R.~Salakhutdinov,
  ``Learning factorized multimodal representations,'' \emph{In International
  Conference on Representation Learning.}, 2019.

\bibitem{pham2018found}
H.~Pham, P.~P. Liang, T.~Manzini, L.-P. Morency, and B.~Poczos, ``Found in
  translation: Learning robust joint representations by cyclic translations
  between modalities,'' \emph{arXiv preprint arXiv:1812.07809}, 2019.

\bibitem{hazarika2018conversational}
D.~Hazarika, S.~Poria, A.~Zadeh, E.~Cambria, L.-P. Morency, and R.~Zimmermann,
  ``Conversational memory network for emotion recognition in dyadic dialogue
  videos,'' in \emph{Proceedings of the 2018 Conference of the North American
  Chapter of the Association for Computational Linguistics: Human Language
  Technologies, Volume 1 (Long Papers)}, vol.~1, 2018, pp. 2122--2132.

\bibitem{poria2017multi}
S.~Poria, E.~Cambria, D.~Hazarika, N.~Mazumder, A.~Zadeh, and L.-P. Morency,
  ``Multi-level multiple attentions for contextual multimodal sentiment
  analysis,'' in \emph{2017 IEEE International Conference on Data Mining
  (ICDM)}.\hskip 1em plus 0.5em minus 0.4em\relax IEEE, 2017, pp. 1033--1038.

\bibitem{liang2018multimodal}
P.~P. Liang, Z.~Liu, A.~Zadeh, and L.-P. Morency, ``Multimodal language
  analysis with recurrent multistage fusion,'' \emph{arXiv preprint
  arXiv:1808.03920}, 2018.

\bibitem{liu2018efficient}
Z.~Liu, Y.~Shen, V.~B. Lakshminarasimhan, P.~P. Liang, A.~Zadeh, and L.-P.
  Morency, ``Efficient low-rank multimodal fusion with modality-specific
  factors,'' \emph{In Proceedings of the 56th Annual Meeting of the Association
  for Computational Linguistics (Volume 1: Long Papers). 2247–2256.
  https://www.aclweb.org/anthology/P18-1209.pdf}, 2018.

\bibitem{barezi2018modality}
E.~J. Barezi, P.~Momeni, P.~Fung \emph{et~al.}, ``Modality-based factorization
  for multimodal fusion,'' \emph{arXiv preprint arXiv:1811.12624}, 2018.

\bibitem{vaswani2017attention}
A.~Vaswani, N.~Shazeer, N.~Parmar, J.~Uszkoreit, L.~Jones, A.~N. Gomez,
  {\L}.~Kaiser, and I.~Polosukhin, ``Attention is all you need,'' in
  \emph{Advances in neural information processing systems}, 2017, pp.
  5998--6008.

\bibitem{tsai2019multimodal}
Y.-H.~H. Tsai, S.~Bai, P.~P. Liang, J.~Z. Kolter, L.-P. Morency, and
  R.~Salakhutdinov, ``Multimodal transformer for unaligned multimodal language
  sequences,'' \emph{arXiv preprint arXiv:1906.00295}, 2019.

\bibitem{pham2019found}
H.~Pham, P.~P. Liang, T.~Manzini, L.-P. Morency, and B.~P{\'o}czos, ``Found in
  translation: Learning robust joint representations by cyclic translations
  between modalities,'' in \emph{Proceedings of the AAAI Conference on
  Artificial Intelligence}, vol.~33, no.~01, 2019, pp. 6892--6899.

\bibitem{tsai2020multimodal}
Y.-H.~H. Tsai, M.~Q. Ma, M.~Yang, R.~Salakhutdinov, and L.-P. Morency,
  ``Multimodal routing: Improving local and global interpretability of
  multimodal language analysis,'' in \emph{Proceedings of the Conference on
  Empirical Methods in Natural Language Processing. Conference on Empirical
  Methods in Natural Language Processing}, vol. 2020.\hskip 1em plus 0.5em
  minus 0.4em\relax NIH Public Access, 2020, p. 1823.

\bibitem{yang2021mtag}
J.~Yang, Y.~Wang, R.~Yi, Y.~Zhu, A.~Rehman, A.~Zadeh, S.~Poria, and L.-P.
  Morency, ``Mtag: Modal-temporal attention graph for unaligned human
  multimodal language sequences,'' in \emph{Proceedings of the 2021 Conference
  of the North American Chapter of the Association for Computational
  Linguistics: Human Language Technologies}, 2021, pp. 1009--1021.

\bibitem{wollmer2013youtube}
M.~W{\"o}llmer, F.~Weninger, T.~Knaup, B.~Schuller, C.~Sun, K.~Sagae, and L.-P.
  Morency, ``Youtube movie reviews: Sentiment analysis in an audio-visual
  context,'' \emph{IEEE Intelligent Systems}, vol.~28, no.~3, pp. 46--53, 2013.

\bibitem{perez2013utterance}
V.~P{\'e}rez-Rosas, R.~Mihalcea, and L.-P. Morency, ``Utterance-level
  multimodal sentiment analysis,'' in \emph{Proceedings of the 51st Annual
  Meeting of the Association for Computational Linguistics (Volume 1: Long
  Papers)}, 2013, pp. 973--982.

\bibitem{busso2008iemocap}
C.~Busso, M.~Bulut, C.-C. Lee, A.~Kazemzadeh, E.~Mower, S.~Kim, J.~N. Chang,
  S.~Lee, and S.~S. Narayanan, ``Iemocap: Interactive emotional dyadic motion
  capture database,'' \emph{Language resources and evaluation}, vol.~42, no.~4,
  pp. 335--359, 2008.

\bibitem{ekman1997face}
R.~Ekman, \emph{What the face reveals: Basic and applied studies of spontaneous
  expression using the Facial Action Coding System (FACS)}.\hskip 1em plus
  0.5em minus 0.4em\relax Oxford University Press, USA, 1997.

\bibitem{zadeh2017tensor}
A.~Zadeh, M.~Chen, S.~Poria, E.~Cambria, and L.-P. Morency, ``Tensor fusion
  network for multimodal sentiment analysis,'' \emph{In Proceedings of the 2017
  Conference on Empirical Methods in Natural Language Processing. 1103–1114.
  https://www.aclweb.org/anthology/D17-1115.pdf}, 2017.

\bibitem{chen2017multimodal}
M.~Chen, S.~Wang, P.~P. Liang, T.~Baltru{\v{s}}aitis, A.~Zadeh, and L.-P.
  Morency, ``Multimodal sentiment analysis with word-level fusion and
  reinforcement learning,'' in \emph{Proceedings of the 19th ACM International
  Conference on Multimodal Interaction}, 2017, pp. 163--171.

\bibitem{sun2020learning}
Z.~Sun, P.~Sarma, W.~Sethares, and Y.~Liang, ``Learning relationships between
  text, audio, and video via deep canonical correlation for multimodal language
  analysis,'' in \emph{Proceedings of the AAAI Conference on Artificial
  Intelligence}, vol.~34, no.~05, 2020, pp. 8992--8999.

\bibitem{rahman2020integrating}
W.~Rahman, M.~K. Hasan, S.~Lee, A.~B. Zadeh, C.~Mao, L.-P. Morency, and
  E.~Hoque, ``Integrating multimodal information in large pretrained
  transformers,'' in \emph{Proceedings of the 58th Annual Meeting of the
  Association for Computational Linguistics}, 2020, pp. 2359--2369.

\bibitem{sundararajan2017axiomatic}
M.~Sundararajan, A.~Taly, and Q.~Yan, ``Axiomatic attribution for deep
  networks,'' in \emph{Proceedings of the 34th International Conference on
  Machine Learning-Volume 70}.\hskip 1em plus 0.5em minus 0.4em\relax JMLR.
  org, 2017, pp. 3319--3328.

\end{thebibliography}





\end{document}